\documentclass[10pt, conference]{IEEEtran} 
\usepackage[utf8]{inputenc}
\usepackage{amsmath, amssymb, amsfonts}
\usepackage{graphicx}
\usepackage{booktabs}
\usepackage{xcolor}
\usepackage{url}
\usepackage{algorithm}
\usepackage[noend]{algpseudocode}
\usepackage{caption}
\usepackage{subcaption}
\usepackage{hyperref}
\usepackage{fancyhdr} 
\usepackage{cite}
\usepackage{enumitem} 
\usepackage{cuted}      
\usepackage{stfloats}   
\usepackage{tikz}       
\usepackage{bm}
\usepackage{mathtools}

\hypersetup{
colorlinks=true,
linkcolor=blue,
filecolor=magenta,
urlcolor=cyan,
pdftitle={ViT-based Class-conditioned Autoencoder (ViTCAE)},
pdfpagemode=FullScreen,
}

\newcommand{\modelname}{ViTCAE}
\newcommand{\zglobal}{$z_2$} 
\newcommand{\zlocal}{$z_1$}  
\IEEEoverridecommandlockouts 

\title{\modelname{}: ViT-based Class-conditioned Autoencoder}

\author{
  \begin{tabular}{@{}c@{\hspace{1em}}c@{\hspace{1em}}c@{\hspace{1em}}c@{}}
    Vahid Jebraeeli & Hamid Krim & Derya Cansever \\
    ECE Department & ECE Department & ECE Department \\
    NC State University & NC State University & NC State University \\
    Raleigh, USA & Raleigh, USA & Raleigh, USA \\
    vjebrae@ncsu.edu & ahk@ncsu.edu & dhcansev@ncsu.edu
  \end{tabular}
}

\begin{document}
\maketitle

\begin{abstract}
Vision Transformer (ViT) based autoencoders often underutilize the global Class token and employ static attention mechanisms, limiting both generative control and optimization efficiency. This paper introduces \modelname{}, a framework that addresses these issues by re-purposing the Class token into a generative linchpin. In our architecture, the encoder maps the Class token to a global latent variable that dictates the prior distribution for local, patch-level latent variables, establishing a robust dependency where global semantics directly inform the synthesis of local details. Drawing inspiration from opinion dynamics, we treat each attention head as a dynamical system of interacting tokens seeking consensus. This perspective motivates a convergence-aware temperature scheduler that adaptively anneals each head’s influence function based on its distributional stability. This process enables a principled head-freezing mechanism, guided by theoretically-grounded diagnostics like an attention evolution distance and a consensus/cluster functional. This technique prunes converged heads during training to significantly improve computational efficiency without sacrificing fidelity. By unifying a generative Class token with an adaptive attention mechanism rooted in multi-agent consensus theory, \modelname{} offers a more efficient and controllable approach to transformer-based generation. \footnote{Thanks to the generous support of ARO grant W911NF-23-2-0041.}
\end{abstract}


\section{Introduction}
Transformer architectures have reshaped computer vision research by replacing convolutional hierarchies with sequence processing. The Vision Transformer (ViT)~\cite{dosovitskiy2021an}, when coupled with \emph{Masked Autoencoders} (MAE)~\cite{he2022mae}, achieves state-of-the-art accuracy while relaxing the dependence on large-scale labeled data. Yet two assumptions inherited from the original ViT design remain largely unchallenged.

\textbf{First}, the global Class token is traditionally treated as a passive feature pooler. MAE simply prepends a single learnable vector to the patch sequence and expects it to absorb image semantics without any explicit generative role. Recent diagnostic studies reveal that this bottleneck hinders optimization and leads to uneven reconstruction quality across spatial regions.

\textbf{Second}, the entropy of attention heads is governed by a fixed softmax temperature. This temperature dictates the selectivity of each head; ill-matched values can slow convergence or trigger gradient instabilities. While manual temperature adjustments have proved useful in other domains, adaptive control has yet to enter mainstream ViT practice.

Beyond these vision-centric issues, we draw inspiration from the opinion dynamics literature~\cite{motsch2014heterophilious, jabin2014clustering}, which models how distributed agents converge toward consensus or form distinct clusters based on interaction rules. By viewing the Vision Transformer as a multi-agent system, where tokens are agents and the self-attention mechanism acts as a learnable influence function~\cite{jabin2014clustering}, we can analyze its layer-wise processing as a dynamic evolution. This perspective suggests that adaptively modulating interaction strengths (attention temperatures) can more efficiently guide the system toward stable, meaningful representations. We therefore revisit the static interaction rules inside ViT attention blocks.

This pursuit of more efficient and controllable model dynamics builds upon our prior work on dataset manipulation. We previously explored data-centric approaches, developing frameworks for both dataset condensation~\cite{jebraeeli2024koopcon}, to distill large datasets into compact, information-rich cores, and expansive synthesis~\cite{jebraeeli2025generative}, to generate extensive datasets from minimal samples. While these methods effectively addressed data scarcity and abundance, they underscored the limitations of existing generative backbones. This motivated a shift in focus from manipulating the dataset itself to re-engineering the fundamental generative architecture for superior control and efficiency in creating new synthetic datapoints. Our current work, therefore, pushes beyond data-level synthesis to refine the internal mechanics of the generative model itself.

In this work, we introduce the \emph{ViT-based Class-conditioned Autoencoder} (ViTCAE), a self-supervised model that tackles the limitations above in a unified framework. ViTCAE endows the Class token with a \emph{generative} mandate: the encoder first predicts a compact latent vector $z$ from the token and, conditioned on $z$, infers a learnable prior over patch tokens; a lightweight decoder then reconstructs the image from these token embeddings. To actively control the system's internal dynamics, every attention head is equipped with a learnable temperature $\tau_{l,h}$ that is \emph{automatically annealed} according to the Wasserstein drift of its attention distribution. This adaptively reshapes each head's influence function, enforcing a coarse-to-fine learning curriculum. Once a head's temperature and output entropy stabilise, its parameters are \emph{frozen}, reducing both training and inference FLOPs without compromising accuracy.

\medskip
Here are our contributions:
\begin{itemize}[leftmargin=2em]
    \item We propose ViTCAE, the first auto-encoding framework that conditions patch reconstruction on a generative Class latent, thereby unifying global semantics and local detail within a single ViT backbone.
    \medskip
    \item We devise a convergence-aware, per-head temperature scheduler that modulates each head's influence based on an optimal-transport drift of its attention, and pair it with two diagnostics of head convergence grounded in opinion dynamics theory, a transport-based evolution distance and a consensus/cluster functional, yielding more stable optimisation and principled head freezing.
    \medskip
    \item We introduce a head-freezing mechanism that automatically detects converged heads, achieving up to 25\,\% FLOP savings with negligible loss in reconstruction fidelity.
\end{itemize}

\medskip
Chapter organization is as bellow. Section~\ref{sec:background} reviews related work on MAE, CLS-token design, adaptive attention, opinion dynamics, and head compression. Section~\ref{sec:method} details the ViTCAE architecture and the temperature-control algorithm. Comprehensive experiments and ablations are presented in Section~\ref{sec:experiments}, followed by conclusions in Section~\ref{sec:conclusion}.

\section{Related Background}\label{sec:background}

Our work, ViTCAE, integrates concepts from several domains, re-framing the Vision Transformer architecture through the lens of dynamical systems and opinion dynamics. To motivate our approach, this section provides the necessary background on the core ViT architecture, its interpretation as a dynamic system, the theory of opinion dynamics that inspires our adaptive mechanisms, and the generative modeling techniques we employ~\cite{jiang2021transgan,xian2021vitvae}.

\subsection{Vision Transformer Architecture}\label{sec:background:vit_arch}
The Vision Transformer (ViT)~\cite{dosovitskiy2021an} marks a significant paradigm shift in computer vision by demonstrating that a pure transformer architecture can achieve state-of-the-art results on image recognition tasks, challenging the long-standing dominance of Convolutional Neural Networks (CNNs). The core principle of ViT is to remodel an image as a sequence of patches and process it using the standard Transformer encoder originally developed for natural language processing~\cite{vaswani2017attention}. This process involves three main stages: patch embedding, sequential processing through transformer blocks, and final representation extraction.

\begin{figure*}[t]
    \centering
    \includegraphics[width=0.9\textwidth]{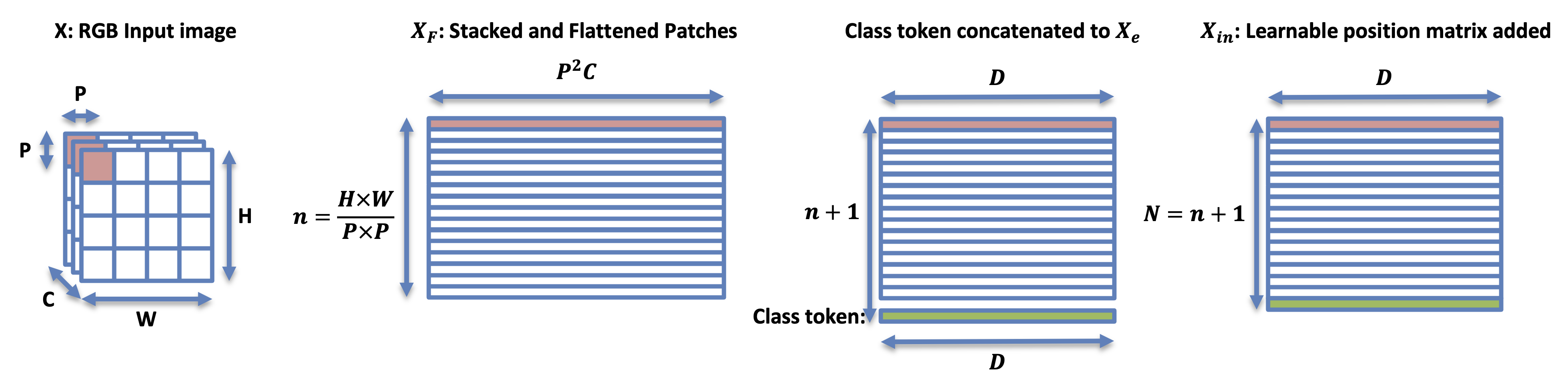}
    \caption{The ViT embedding process. An input image is divided into patches, which are flattened and linearly projected into patch embeddings. A learnable class token is prepended, and positional embeddings are added to the entire sequence to create the final input for the Transformer.}
    \label{fig:vit_embedding}
\end{figure*}

\paragraph{Patching and Embedding}
The initial step transforms a 2D image into a 1D sequence of token embeddings, making it suitable for a Transformer, as illustrated in Figure~\ref{fig:vit_embedding}. Given an input image $\bm{X} \in \mathbb{R}^{H \times W \times C}$, where $H$, $W$, and $C$ are the height, width, and number of channels, respectively, the process is as follows:

\begin{enumerate}
    \item \textbf{Patchification:} The image $\bm{X}$ is reshaped into a series of flattened 2D patches. It is divided into $n$ non-overlapping patches, each of size $P \times P$. The total number of patches is given by:
    \begin{equation}
        n = \frac{H \times W}{P^2}.
    \end{equation}
    Each patch is flattened into a vector, resulting in a sequence of $n$ patch vectors. These are stacked to form a matrix of flattened patches, $\bm{X}_F \in \mathbb{R}^{n \times (P^2 C)}$.

    \item \textbf{Linear Projection:} To project the patches into a $D$-dimensional embedding space that the Transformer can process, the flattened patches $\bm{X}_F$ are multiplied by a learnable embedding matrix $\bm{E} \in \mathbb{R}^{(P^2 C) \times D}$. This yields the sequence of patch embeddings $\bm{X}_e$:
    \begin{equation}
        \bm{X}_e = \bm{X}_F \bm{E}, \quad \text{where} \quad \bm{X}_e \in \mathbb{R}^{n \times D}.
    \end{equation}

    \item \textbf{Class Token and Positional Embeddings:} In line with the original ViT design for classification, a special learnable embedding, the Class token ($\text{class token} \in \mathbb{R}^{1 \times D}$), is prepended to the sequence of patch embeddings. This token is designed to aggregate global information from the entire sequence. To retain spatial information, which is lost during patchification, learnable 1D position embeddings ($\bm{pos} \in \mathbb{R}^{(n+1) \times D}$) are added to the token embeddings~\cite{su2021roformer}. The final input sequence for the Transformer encoder, $\bm{X}_{in}$, is constructed as:
    \begin{equation}
        \bm{X}_{in} = [\bm{X}_e; \text{class token}] + \bm{pos}, \, \text{where} \, \bm{X}_{in} \in \mathbb{R}^{(n+1) \times D}.
    \end{equation}
    For convenience, we denote the sequence length as $N = n+1$.
\end{enumerate}

\paragraph{Transformer Block}
The input sequence $\bm{X}_{in}$ is then passed through a stack of $L$ identical Transformer blocks. Each block is composed of two main sub-layers: a Multi-Head Self-Attention (MHSA) module and a position-wise Feed-Forward Network (FFN).

\begin{figure*}[t]
    \centering
    \includegraphics[width=\textwidth]{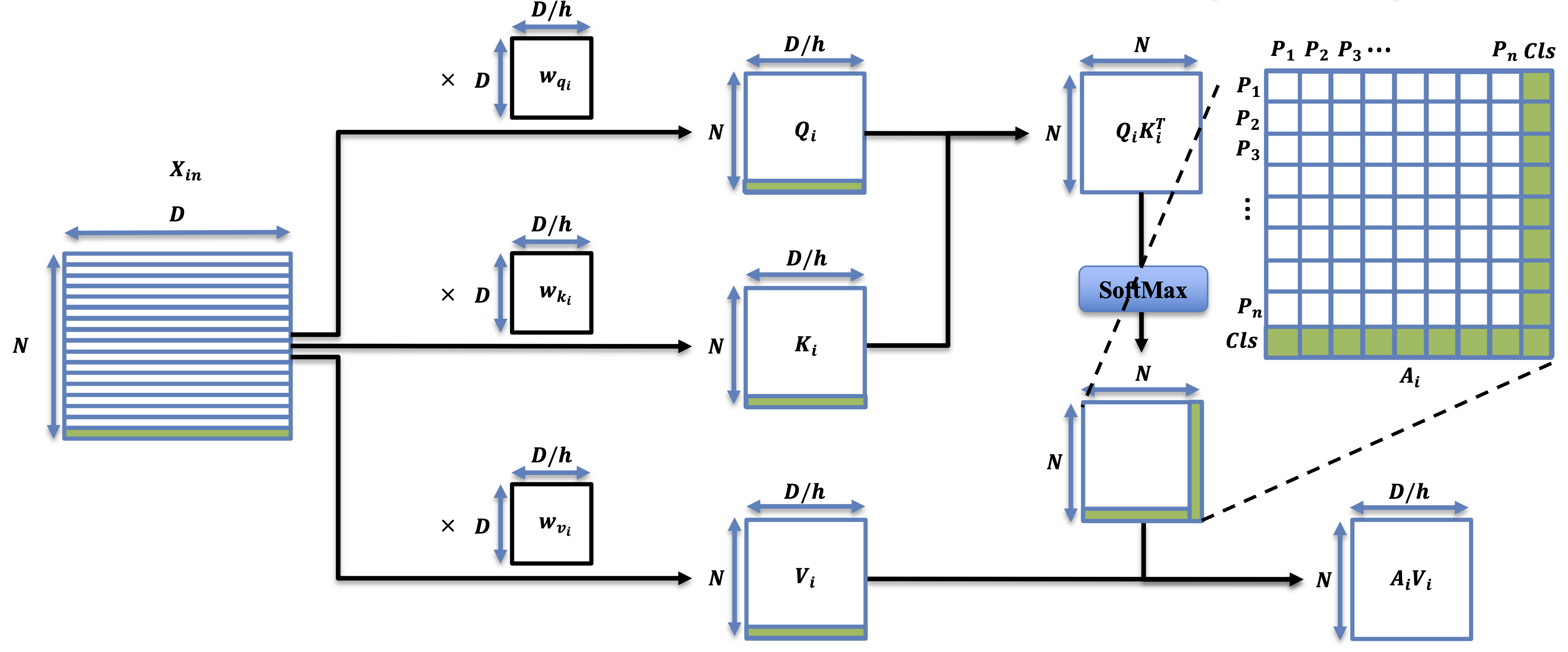}
    \caption{Data flow within a single attention head. The input sequence $\bm{X}_{in}$ is projected into Query (Q), Key (K), and Value (V) matrices. The dot product of Q and K, followed by a softmax, yields the attention matrix $\bm{A}_i$, which weights the V matrix to produce the head's output.}
    \label{fig:vit_attention}
\end{figure*}

The core of the Transformer is the self-attention mechanism, which allows tokens to interact and aggregate information from across the entire sequence. The MHSA module runs multiple single attention "heads" in parallel. The mechanism of a single attention head is shown in Figure~\ref{fig:vit_attention}. For each head $i$, the input sequence $\bm{X}_{in} \in \mathbb{R}^{N \times D}$ is linearly projected into Query ($\bm{Q}_i$), Key ($\bm{K}_i$), and Value ($\bm{V}_i$) matrices using learnable weight matrices $\bm{W}_{q_i}, \bm{W}_{k_i}, \bm{W}_{v_i} \in \mathbb{R}^{D \times (D/h)}$, where $h$ is the total number of heads. For simplicity, the dimension of the keys, queries, and values for each head, $D/h$, will be denoted as $d_k$ in subsequent sections.
\begin{align}
    \bm{Q}_i &= \bm{X}_{in} \bm{W}_{q_i} \\
    \bm{K}_i &= \bm{X}_{in} \bm{W}_{k_i} \\
    \bm{V}_i &= \bm{X}_{in} \bm{W}_{v_i}.
\end{align}
The attention scores are computed via scaled dot-product between queries and keys, which are then used to create a weighted sum of the values. The output of a single attention head is:
\begin{equation}
    \text{Head}_i = \text{softmax}\left(\frac{\bm{Q}_i \bm{K}_i^T}{\sqrt{d_k}}\right) \bm{V}_i = \bm{A}_i \bm{V}_i,
\end{equation}
where $\bm{A}_i \in \mathbb{R}^{N \times N}$ is the attention matrix for head $i$. 

\begin{figure*}[t]
    \centering
    \includegraphics[width=\textwidth]{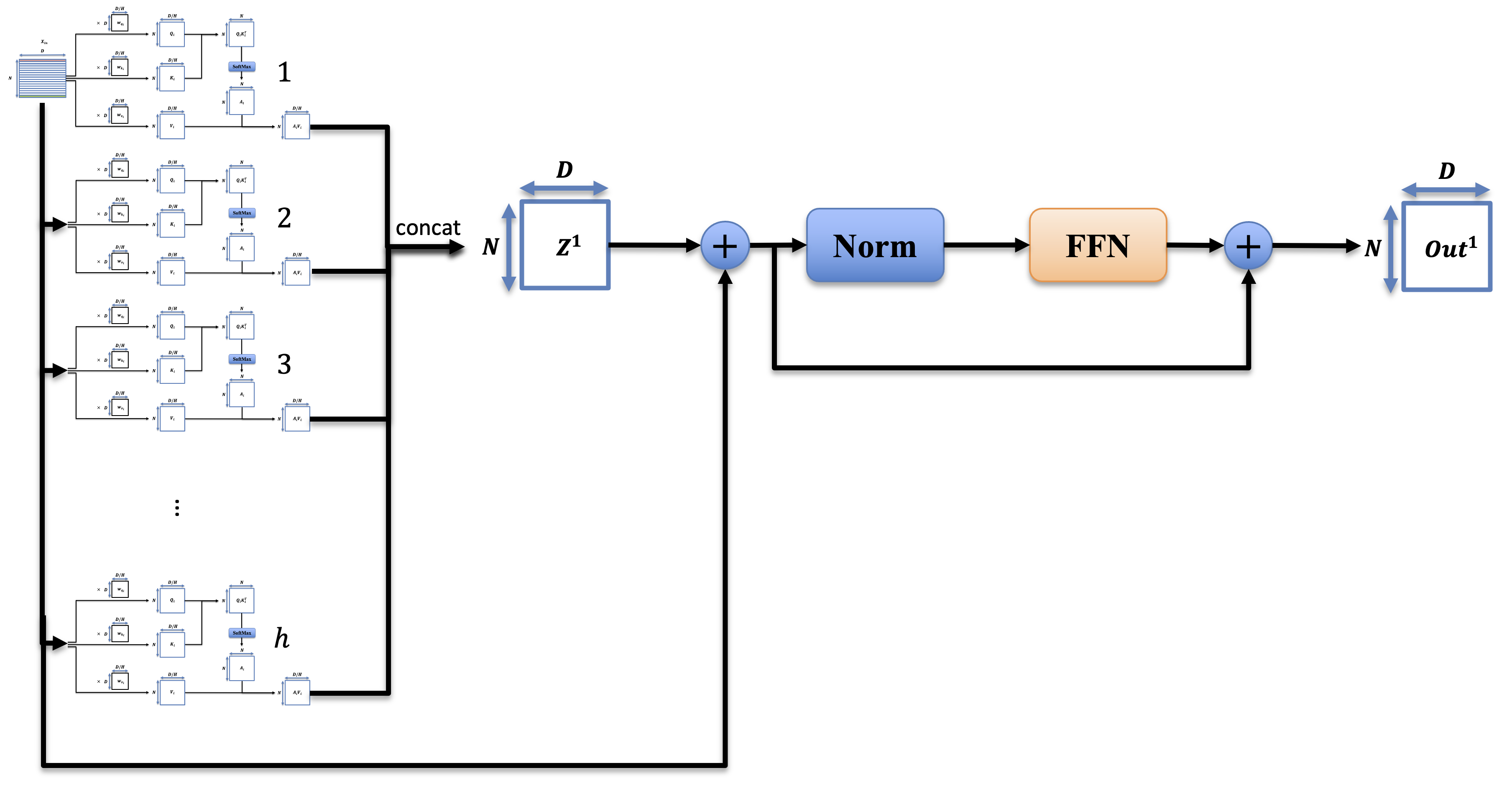}
    \caption{Structure of a Transformer Block/Layer. It consists of multiple attention heads running in parallel whose outputs are concatenated. This Multi-Head Self-Attention sub-layer and a subsequent Feed-Forward Network sub-layer both feature residual connections and layer normalization.}
    \label{fig:vit_block}
\end{figure*}

\begin{figure*}[t]
    \centering
    \includegraphics[width=\textwidth]{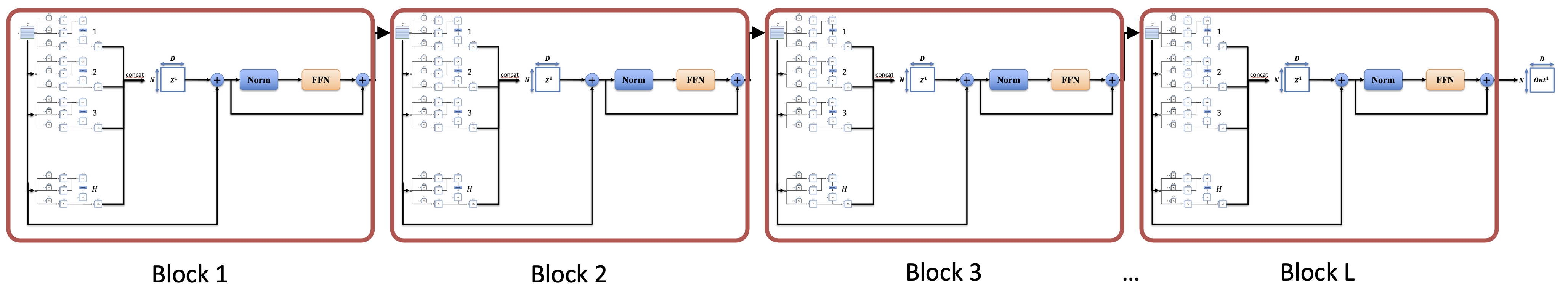}
    \caption{A complete Vision Transformer Encoder Network. The architecture consists of $L$ identical Transformer blocks stacked sequentially. The output of one block serves as the input to the next.}
    \label{fig:vit_network}
\end{figure*}

As shown in Figure~\ref{fig:vit_block}, the outputs of all $h$ heads are concatenated and passed through a final linear layer to produce the output of the MHSA module, $\bm{Z}^l \in \mathbb{R}^{N \times D}$. Each Transformer layer applies Layer Normalization before each sub-layer and employs residual connections around them. The full operation of a layer $l$ on an input $\bm{X}^{l-1}$ can be summarized as:
\begin{align}
    \bm{X}' &= \text{MHSA}(\text{Norm}(\bm{X}^{l-1})) + \bm{X}^{l-1} \\
    \bm{X}^{l} &= \text{FFN}(\text{Norm}(\bm{X}')) + \bm{X}'.
\end{align}
This entire structure is repeated $L$ times to form the complete Transformer network, as depicted in Figure~\ref{fig:vit_network}. The output embedding corresponding to the Class token from the final layer is often used as the aggregate representation of the image for downstream tasks.

\subsection{Vision Transformers as a Dynamical System}\label{sec:background:vit_dynamics}
The deep, sequential nature of the ViT architecture allows the layer-wise processing to be interpreted as the discretization of an Ordinary Differential Equation (ODE)~\cite{weinan2017proposal, chen2018neural}. This viewpoint interprets the layer-wise processing of token embeddings as a discretization of an ODE, providing a continuous-depth model of the network's behavior where the token embeddings $\bm{x}(t)$ evolve over a continuous depth variable $t \in [0, L]$:
\begin{equation}
    \frac{d\bm{x}(t)}{dt} = F(\bm{x}(t), t),
\end{equation}
where $F$ is the function learned by the block's parameters. This continuous-depth viewpoint provides a powerful tool for analyzing information flow and motivates a deeper look into the nature of the interaction function $F$ defined by the self-attention mechanism.

\subsection{Opinion Dynamics as a Model for Self-Attention}\label{sec:background:opinion_dynamics}
Opinion dynamics models how a group of interacting agents reaches a collective state, such as consensus or polarization. These models provide a powerful analogy for understanding the self-attention mechanism. The canonical continuous-time model describes the evolution of each agent's "opinion" (a state vector) $x_i$ as an alignment with its neighbors:
\begin{equation} \label{eq:opinion_model}
    \dot{x}_{i} = \frac{\sum_{j}\phi_{ij}(x_{j}-x_{i})}{\sum_{j}\phi_{ij}}, \quad \text{where} \quad \phi_{ij} = \phi(|x_{j}-x_{i}|).
\end{equation}
Here, $\phi$ is the influence function, which quantifies the interaction strength between agents based on the difference between their states. The system evolves until reaching a stable equilibrium, which can be a single global consensus or a set of distinct clusters. The properties of $\phi$ are critical; for instance, "heterophilious" dynamics (where $\phi$ increases with distance) can enhance global consensus.

We map this framework to the Vision Transformer as follows:
\begin{itemize}[leftmargin=1.5em]
    \item \textbf{Agents and Opinions:} Tokens in a sequence are agents; their embedding vectors are their opinions.
    \item \textbf{Influence Function:} The softmax-normalized attention scores act as the influence function, where the attention weight $A_{ij}$ is the influence of token $j$ on token $i$.
    \item \textbf{Dynamic Evolution:} The sequential application of Transformer layers corresponds to the time evolution of the system.
    \item \textbf{Consensus and Clustering:} A converged attention head represents a stable state where tokens have formed fixed relational patterns, analogous to agents reaching a consensus or forming stable clusters. Our work leverages this view to diagnose and control head convergence.
\end{itemize}

\subsection{Generative Modeling Components}\label{sec:background:components}
Our architecture integrates several key generative modeling techniques.

\paragraph{Variational Autoencoders}
The model is a hierarchical Variational Autoencoder (VAE)~\cite{kingma2013auto,sonderby2016ladder,vahdat2020nvae,kim2018disentangling}. A VAE consists of an encoder (inference network) $q_{\phi}(z|x)$ and a decoder (generative network) $p_{\theta}(x|z)$. The encoder approximates the true but intractable posterior $p(z|x)$ by mapping an input $x$ to the parameters of a distribution in the latent space, typically a diagonal Gaussian: $q_{\phi}(z|x) = \mathcal{N}(\mu_{\phi}(x), \operatorname{diag}\sigma^2_{\phi}(x))$. To generate a sample $\hat{z}$, the reparameterization trick is used:
\begin{equation}
    \hat{z} = \mu_{\phi}(x) + \sigma_{\phi}(x) \odot \epsilon, \quad \text{where} \quad \epsilon \sim \mathcal{N}(0, I).
\end{equation}
The VAE is trained by maximizing the Evidence Lower Bound (ELBO):
\begin{equation}\label{eq:elbo}
    \mathcal{L}_{\text{ELBO}} = \mathbb{E}_{q_{\phi}(z|x)}[\log p_{\theta}(x|z)] - \mathrm{D}_{\mathrm{KL}}(q_{\phi}(z|x) \,\|\, p(z)).
\end{equation}
This objective balances reconstruction fidelity with a Kullback-Leibler (KL) divergence term that regularizes the latent space~\cite{higgins2017beta} by encouraging the approximate posterior to match a prior $p(z)$, commonly $\mathcal{N}(0, I)$. Our model employs a conditional version of this framework, as clarified in Section~\ref{sec:method:loss}.

\paragraph{Wasserstein-based Regularization}
To overcome the limitations of the KL divergence, especially after an initial training phase, we regularize the latent space using a Wasserstein-type metric. This is implemented via the Maximum Mean Discrepancy (MMD)~\cite{gretton2012kernel} with a characteristic Inverse-Multiquadratic (IMQ) kernel~\cite{tolstikhin2018wasserstein}:
\begin{equation}
    k_{\mathrm{IMQ}}(x, y) = \frac{C}{C + \|x-y\|_2^2}.
\end{equation}
Given a kernel function $k(\cdot, \cdot)$, the squared MMD between samples from distributions $P_r$ and $P_g$ can be computed as:
\begin{equation}
\begin{aligned}
\operatorname{MMD}^2(P_r, P_g)
&= \mathbb{E}_{x, x' \sim P_r}[k(x, x')]
    - 2\,\mathbb{E}_{x \sim P_r,\, y \sim P_g}[k(x, y)] \\
&\quad + \mathbb{E}_{y, y' \sim P_g}[k(y, y')].
\end{aligned}
\end{equation}

\paragraph{Conditional Normalization}
To enable fine-grained control over the generative process, the decoder is conditioned~\cite{sohn2015learning} on the global latent variable $z_{cls}$ using Feature-wise Linear Modulation (FiLM)~\cite{perez2018film}. For an intermediate feature map $\bm{h}$, FiLM applies a feature-wise affine transformation:
\begin{equation}
    \mathrm{FiLM}(\bm{h} | z_{cls}) = \gamma \odot \bm{h} + \beta,
\end{equation}
where the scale $\gamma$ and shift $\beta$ are predicted from $z_{cls}$ by a small MLP.


\section{Methodology}\label{sec:method}

\begin{figure*}[t]
    \centering
    \includegraphics[width=\textwidth]{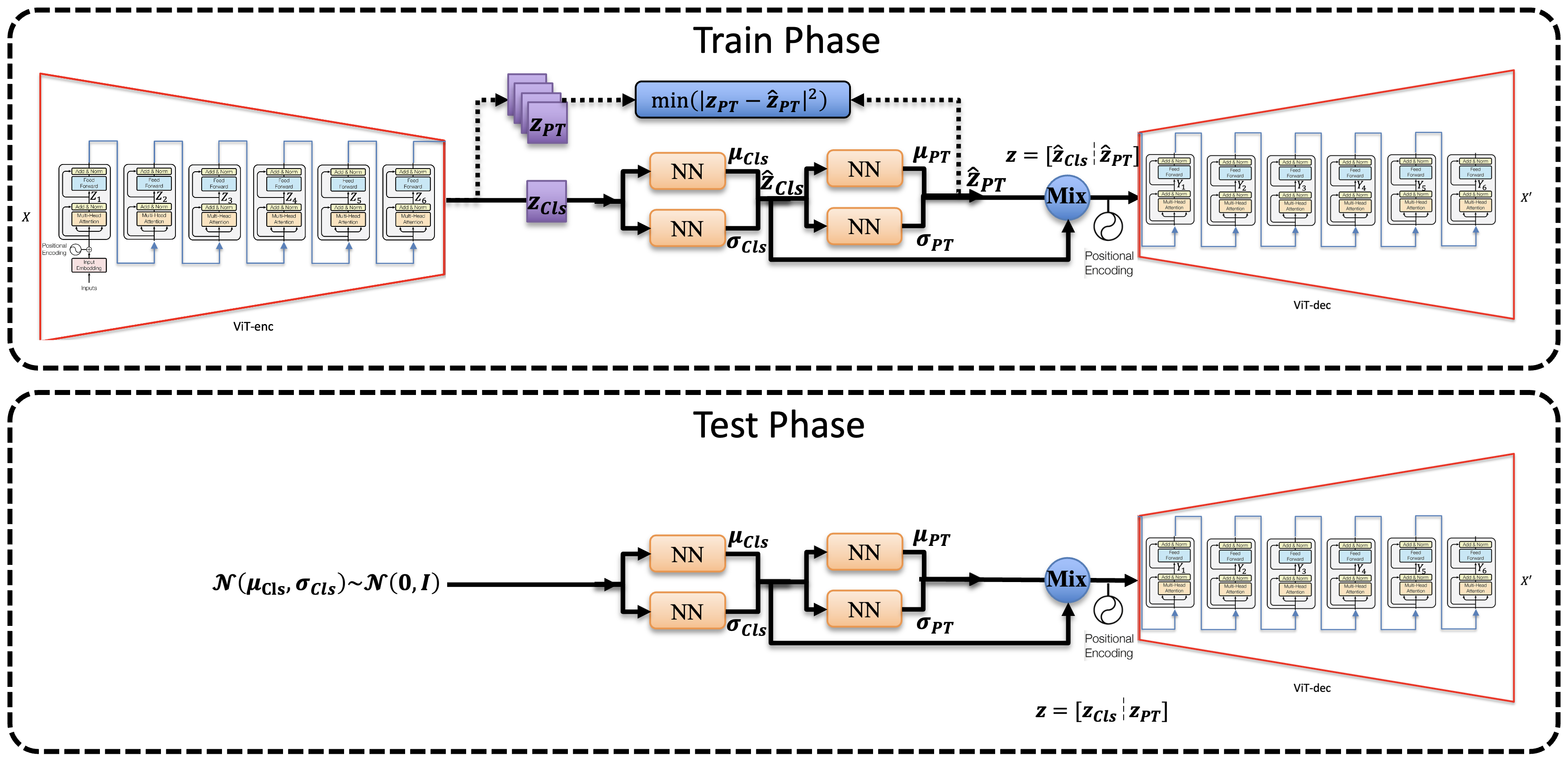}
    \caption{ViTCAE architecture. \textbf{Top (Training):} The encoder produces a global latent \(z_{cls}\) and patch latents \(Z_{\mathrm{PT}}\). A neural prior and a deterministic estimator impose regularization. \textbf{Bottom (Inference):} A sample from \(\mathcal{N}(0,I)\), concatenated with conditionally generated patch tokens \(\tilde Z_{\mathrm{PT}}\), drives the decoder to synthesize \(x'\).}
    \label{fig:architecture}
\end{figure*}

\subsection{Notation and Overall Structure}\label{sec:method:overview}
An input image \(x\in\mathbb{R}^{C\times H\times W}\) is divided into \(N_{\mathrm{pt}}=(H/P)^{2}\) patch vectors. After a linear projection and the addition of positional embeddings, the sequence is prepended with a class token \(\bm c_{0}\) and processed by an $L$-layer ViT encoder \(\mathcal{E}_{\phi}\) to yield a global latent \(z_{cls}\in\mathbb{R}^{d}\) and patch latents \(Z_{\mathrm{PT}}\in\mathbb{R}^{N_{\mathrm{pt}}\times d}\).
\begin{equation}
\bigl[\bm c_{0};\bm p_{1};\dots;\bm p_{N_{\mathrm{pt}}}\bigr]
\;\xmapsto{\ \mathcal{E}_{\phi}\ }\;
\bigl[\bm z_{cls};\bm Z_{\mathrm{PT}}\bigr].
\end{equation}
\medskip
\noindent\textbf{Latent Posterior.}
The encoder outputs parameterize diagonal Gaussian posteriors for the global and patch latents,
\begin{equation}\small
\begin{aligned}
    q_{\phi}(z_{cls}\mid x)
        &=\mathcal{N}\!\bigl(\mu_{cls}(x),\operatorname{diag}\sigma^{2}_{cls}(x)\bigr),\\
    q_{\phi}(z_{\mathrm{PT}}\mid x)
        &=\mathcal{N}\!\bigl(\mu_{\mathrm{PT}}(x),\operatorname{diag}\sigma^{2}_{\mathrm{PT}}(x)\bigr),
\end{aligned}
\end{equation}
via two-layer MLP heads. Reparameterized~\cite{rezende2014stochastic} samples are denoted \(\hat z_{cls}\) and \(\hat Z_{\mathrm{PT}}\).

\medskip
\noindent\textbf{Decoder.}
The ViT decoder \(\mathcal{D}_{\theta}\) receives the concatenated sequence \([\hat z_{cls};\tilde Z_{\mathrm{PT}}]\) and outputs the reconstructed image \(x'\) after a final linear projection and patch re-assembly (Fig.~\ref{fig:architecture}).

\subsection{Encoder Block with Adaptive Influence Function}
For each encoder layer \(\ell\) and head \(h\), a learnable inverse temperature \(\tau_{\ell,h}\) controls the sharpness of the attention mechanism. High \(\tau_{\ell,h}\) (low temperature) creates a sharp, selective influence~\cite{kobayashi2023scalar}, while low \(\tau_{\ell,h}\) leads to a diffuse one.

\begin{equation}
\mathrm{Attn}_{\tau_{\ell,h}}(\bm Q,\bm K,\bm V) = \operatorname{softmax}\!\bigl(\tfrac{\bm Q\bm K^{\!\top}}{\tau_{\ell,h} \sqrt{d_k}}\bigr)\bm V.
\end{equation}
At each epoch \(t\), we compute the $1$-Wasserstein distance \(d^{(t)}_{\ell,h}\) between the CLS-to-patch attention distributions of consecutive epochs. The temperature is updated via the schedule
\begin{equation}\label{eq:tau-update}
    \tau_{\ell,h}^{(t+1)}
        =\Bigl(1+\alpha\,d^{(t)}_{\ell,h}\Bigr)^{-1},
        \qquad\alpha>0,
\end{equation}
which guides the head's dynamics from exploration to exploitation. When its temperature and attention drift stabilize below a threshold, a head \((\ell,h)\) is frozen (\(\nabla_{\phi}=0\)).

\subsection{Neural Prior and Patch–Token Estimator}
A conditional Gaussian prior \(p_{\theta}(z_{\mathrm{PT}}\mid z_{cls})=\mathcal{N}\!\bigl(\hat\mu_{\mathrm{PT}},\operatorname{diag}\hat\sigma^{2}_{\mathrm{PT}}\bigr)\) is produced by an MLP \(f_{\theta}(z_{cls})\). Independently, a second MLP estimator, \(\tilde Z_{\mathrm{PT}}=g_{\theta}(z_{cls})\), maps the same global latent to a deterministic patch matrix.

\subsection{Decoder with FiLM Conditioning}\label{sec:method:decoder}
To guide the reconstruction, the global latent \(z_{cls}\) conditions each decoder layer. This is achieved via (i) cross-attention, where patch tokens attend to \(\hat z_{cls}\), and (ii) a FiLM-modulated feed-forward network, \(\mathrm{FFN}(\bm h)\mapsto \gamma(z_{cls})\odot\mathrm{FFN}(\bm h)+\beta(z_{cls})\). This allows \(z_{cls}\) to influence both where and how information flows~\cite{sohn2015learning}.

\subsection{Per–Head Convergence Diagnostics}
\label{sec:method:convdiag}
For layer $\ell$ and head $h$, let $A^{(t)}_{\ell,h}$ be the attention matrix at epoch $t$. We monitor convergence with two diagnostics.

\paragraph{(i) Attention Evolution Distance (Optimal Transport)}
We measure the change in the \texttt{CLS}-to-patch allocation over training using the $1$-Wasserstein distance between consecutive epochs. The drift of head $(\ell,h)$ at epoch $t$ is
\begin{equation}
\label{eq:emd-drift}
d^{(t)}_{\ell,h}
:= W_{1}\!\bigl(a^{(t-1)}_{\ell,h}\!\restriction_{\mathcal{P}},\;
             a^{(t)}_{\ell,h}\!\restriction_{\mathcal{P}}\bigr),
\end{equation}
where $a^{(t)}_{\ell,h}$ is the CLS token's attention row and $\mathcal{P}$ is the set of patch tokens. Small $d^{(t)}_{\ell,h}$ indicates the head’s focus is stationary. This is the same quantity used in the temperature scheduler~\eqref{eq:tau-update}.

\paragraph{(ii) Consensus/Cluster Functional (Invariant Subspace)}
We assess the consensus structure of a head by forming a random-walk operator $P^{(t)}_{\ell,h}$ from its symmetrized attention matrix. The rank of its long-time limit,
\begin{equation}
\label{eq:kappa}
\kappa^{(t)}_{\ell,h} \;:=\; \operatorname{rank}\bigl(\lim_{m\to\infty}(P^{(t)}_{\ell,h})^{m}\bigr),
\end{equation}
is a soft number of consensus clusters. This quantity, from opinion dynamics theory, corresponds to the number of stable groups in the network of tokens.

\paragraph{Definition (Converged Head)}
Head $(\ell,h)$ is converged if its attention evolution distance becomes stationary and its consensus structure stabilizes, i.e., $d^{(t)}_{\ell,h}\to 0$ and $\kappa^{(t)}_{\ell,h}$ becomes constant.

\subsection{Objective and KL-to-Wasserstein Curriculum}
\label{sec:method:loss}
Our training objective uses a curriculum to ensure stability. Initially, posteriors can be poorly scaled, so a KL divergence term provides well-conditioned gradients to rapidly calibrate the latents. Once the latents become more informative, the KL term is fragile to support mismatch. We thus transition to a mixture that includes a Wasserstein-type discrepancy (implemented via IMQ–MMD), which provides smoother gradients by emphasizing geometric alignment. A residual KL term is retained to prevent variance inflation.

\medskip
The full objective combines reconstruction, patch-token discrepancy, and the scheduled regularization. Let $T_w$ be the warm-up period.
\begin{equation}
\begin{aligned}
\mathcal{L}_{\mathrm{rec}} &= \lambda_{L_1}\|x-x'\|_{1} + \lambda_{L_2}\|x-x'\|_{2}^{2}, \\
\mathcal{L}_{\mathrm{PT}} &= \lambda_{\mathrm{PT}}\|\hat Z_{\mathrm{PT}}-\tilde Z_{\mathrm{PT}}\|_2^2.
\end{aligned}
\end{equation}
The regularization terms for global ($z_{cls}$) and patch ($z_{\mathrm{PT}}$) latents are:
\begin{equation}
\begin{aligned}
\mathrm{KL}_{cls} &= \mathrm{D}_{\mathrm{KL}}\!\big(q_{\phi}(z_{cls}\!\mid x)\,\|\,\mathcal{N}(0,I)\big),\\
\mathrm{KL}_{pt}  &= \mathrm{D}_{\mathrm{KL}}\!\big(q_{\phi}(z_{\mathrm{PT}}\!\mid x)\,\|\,p_{\theta}(\cdot\mid z_{cls})\big).
\end{aligned}
\end{equation}
$W_{cls}$ and $W_{pt}$ denote their Wasserstein (IMQ-MMD) surrogates. The total loss at epoch $t$ is:
\begin{equation}
\label{eq:time_loss}
\mathcal{L}_{\text{total}}(t)=
\left\{
\begin{aligned}
\mathcal{L}_{\mathrm{rec}}
&+\lambda_{cls}\,\mathrm{KL}_{cls}
+\lambda_{pt}\,\mathrm{KL}_{pt}
+\mathcal{L}_{\mathrm{PT}}\\[-0.2em]
&\hspace{1.6em}\text{if } t\le T_w,\\[0.6em]
\mathcal{L}_{\mathrm{rec}}
&+\lambda_{cls}\!\left[\alpha\,W_{cls}+(1-\alpha)\,\mathrm{KL}_{cls}\right]\\
&+\lambda_{pt}\!\left[\alpha\,W_{pt} +(1-\alpha)\,\mathrm{KL}_{pt}\right]
+\mathcal{L}_{\mathrm{PT}}\\[-0.2em]
&\hspace{1.6em}\text{if } t>T_w.
\end{aligned}
\right.
\end{equation}

\subsection{Complexity Reduction via Head Freezing}
We leverage the convergence diagnostics to prune redundant computation. A head $(\ell,h)$ is frozen once it has converged per our definition. Formally, for all epochs $t$ after convergence is detected, we set the gradients of its query, key, value, and temperature weights to zero:
\begin{equation}
\nabla_{\phi}\,W^{\ell,h}_{q,k,v} = 0, \qquad \nabla_{\phi}\,\tau_{\ell,h} = 0.
\end{equation}
The forward pass remains unchanged, but the backward pass is eliminated for that head. Because the attention drift $d^{(t)}_{\ell,h}$ drives both temperature annealing and the convergence check, the freezing mechanism is coherently coupled with the adaptive learning rate of the head's influence function.


\section{Experiments}\label{sec:experiments}

We conducted a series of experiments to evaluate the performance of \modelname{}. Our evaluation focuses on two key areas: (1) a quantitative analysis of the training efficiency and classification performance gained from our adaptive control mechanism, and (2) a qualitative assessment of the model's generative capabilities. For our experiments, we used the CIFAR-10, CelebA ($64 \times 64$)~\cite{liu2015deep}, and Tiny ImageNet datasets. All models were implemented in PyTorch, and key hyperparameters are summarized in Table~\ref{tab:hyperparams}.

\subsection{Analysis of Head Convergence Dynamics}

We begin by validating our core hypothesis: that the proposed adaptive temperature scheduler actively stabilizes the internal dynamics of the attention heads. We use two diagnostics—the attention evolution distance ($d^{(t)}_{\ell,h}$) and the number of consensus clusters ($\kappa^{(t)}_{\ell,h}$)—to track the state of each head throughout training. We compare a baseline ViT without our control mechanism (``Uncontrolled'') against \modelname{} (``Controlled'').

\begin{figure*}[tp]
    \centering
    \begin{subfigure}[b]{\textwidth}
        \centering
        \includegraphics[width=\linewidth]{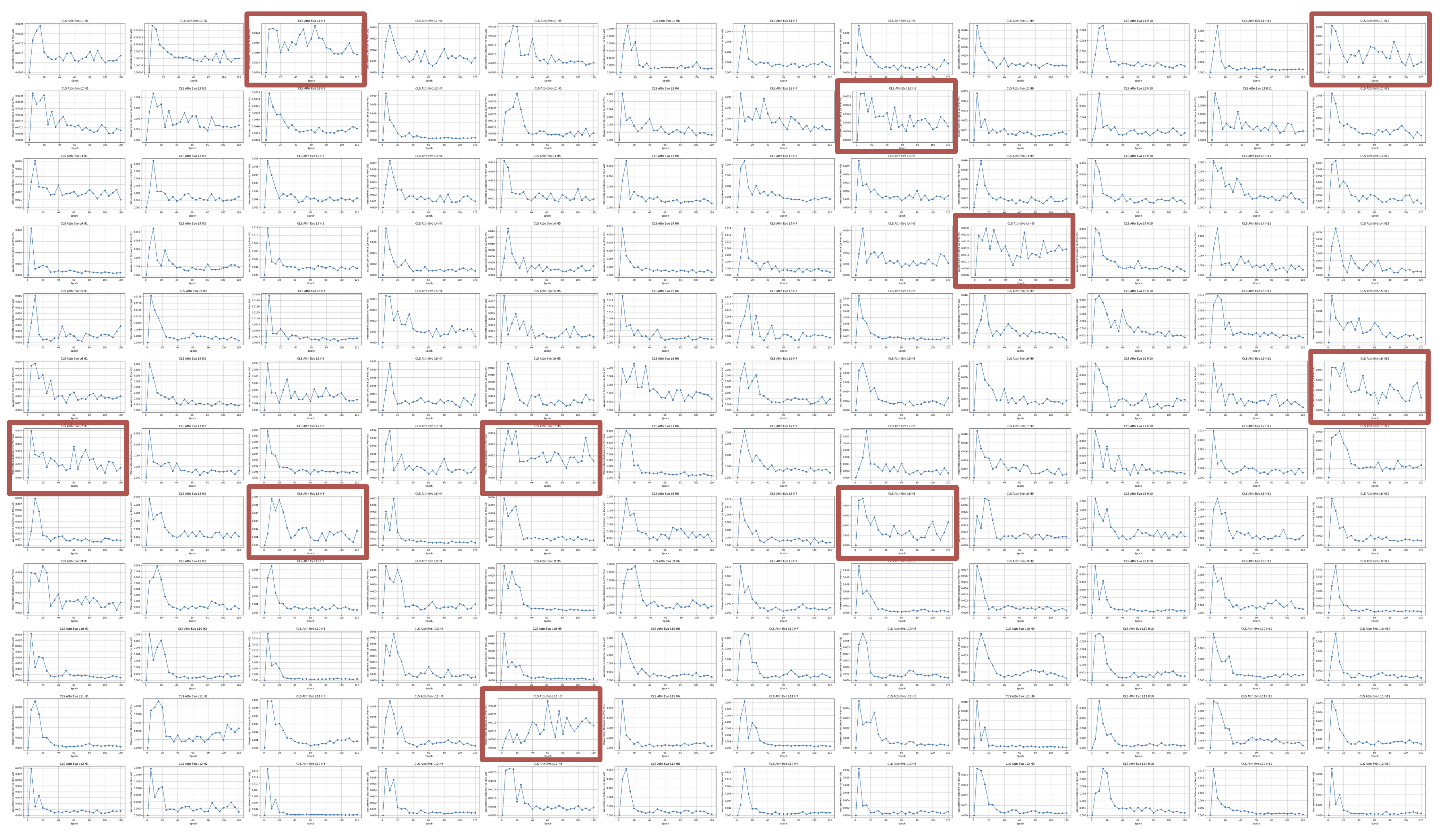}
        \caption{EMD (Attention Evolution) during training without control.}
    \end{subfigure}
    \vfill
    \begin{subfigure}[b]{\textwidth}
        \centering
        \includegraphics[width=\linewidth]{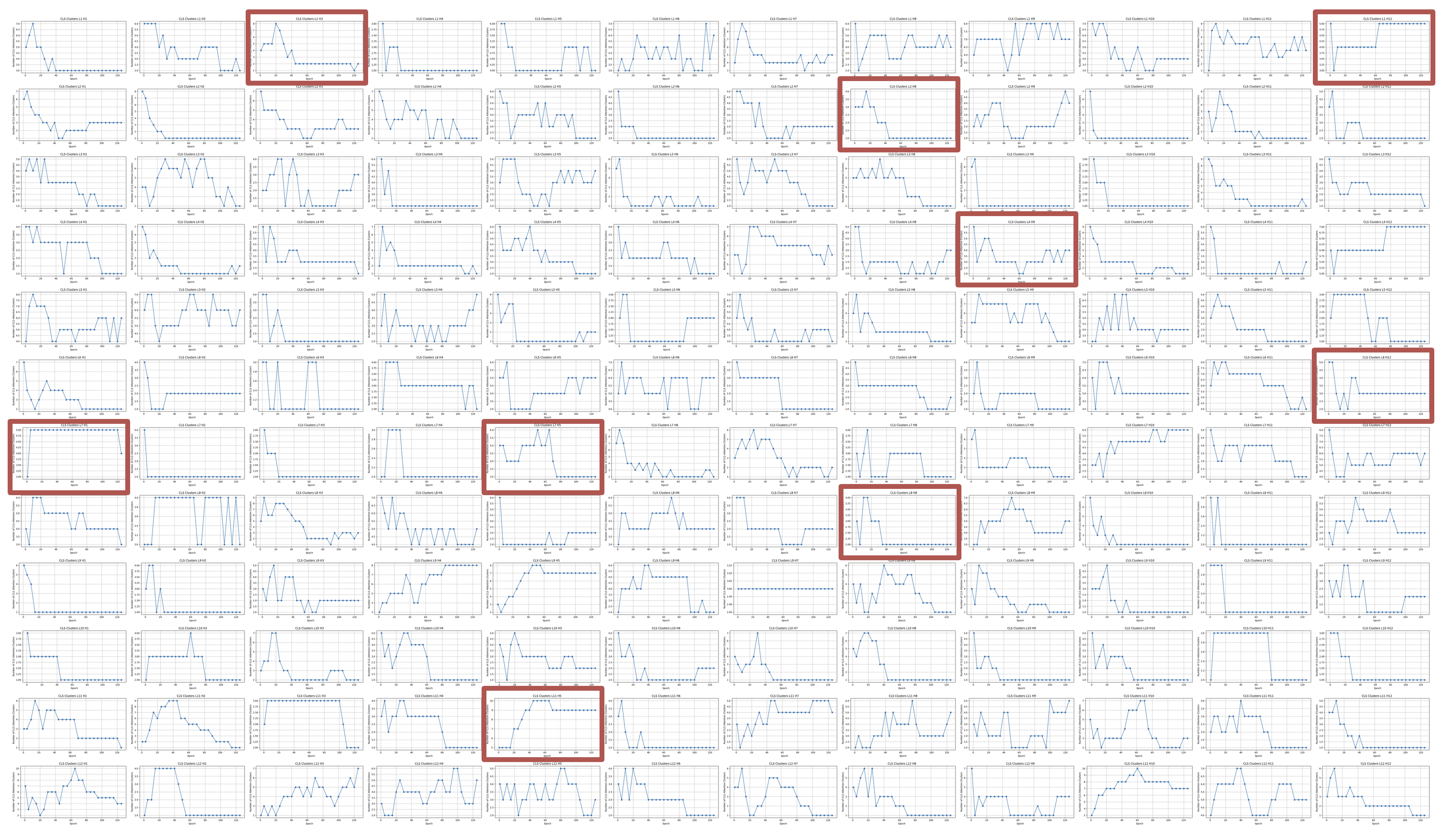}
        \caption{Number of Consensus Clusters during training without control.}
    \end{subfigure}
    \caption{\textbf{Uncontrolled Head Dynamics.} In a baseline model, the per-head attention evolution distance (EMD) and number of consensus clusters ($\kappa$) fluctuate without clear convergence, indicating unstable internal dynamics.}
    \label{fig:dynamics_uncontrolled}
\end{figure*}

Figure~\ref{fig:dynamics_uncontrolled} shows the dynamics of the baseline model. The Earth Mover's Distance (EMD) plots (top) exhibit high variance and no clear trend toward zero, indicating that the attention heads' geometric focus remains unstable. Similarly, the consensus plots (bottom) show the number of clusters fluctuating erratically for many heads, demonstrating a failure to converge to a stable relational structure among tokens.

\begin{figure*}[tp]
    \centering
    \begin{subfigure}[b]{\textwidth}
        \centering
        \includegraphics[width=\linewidth]{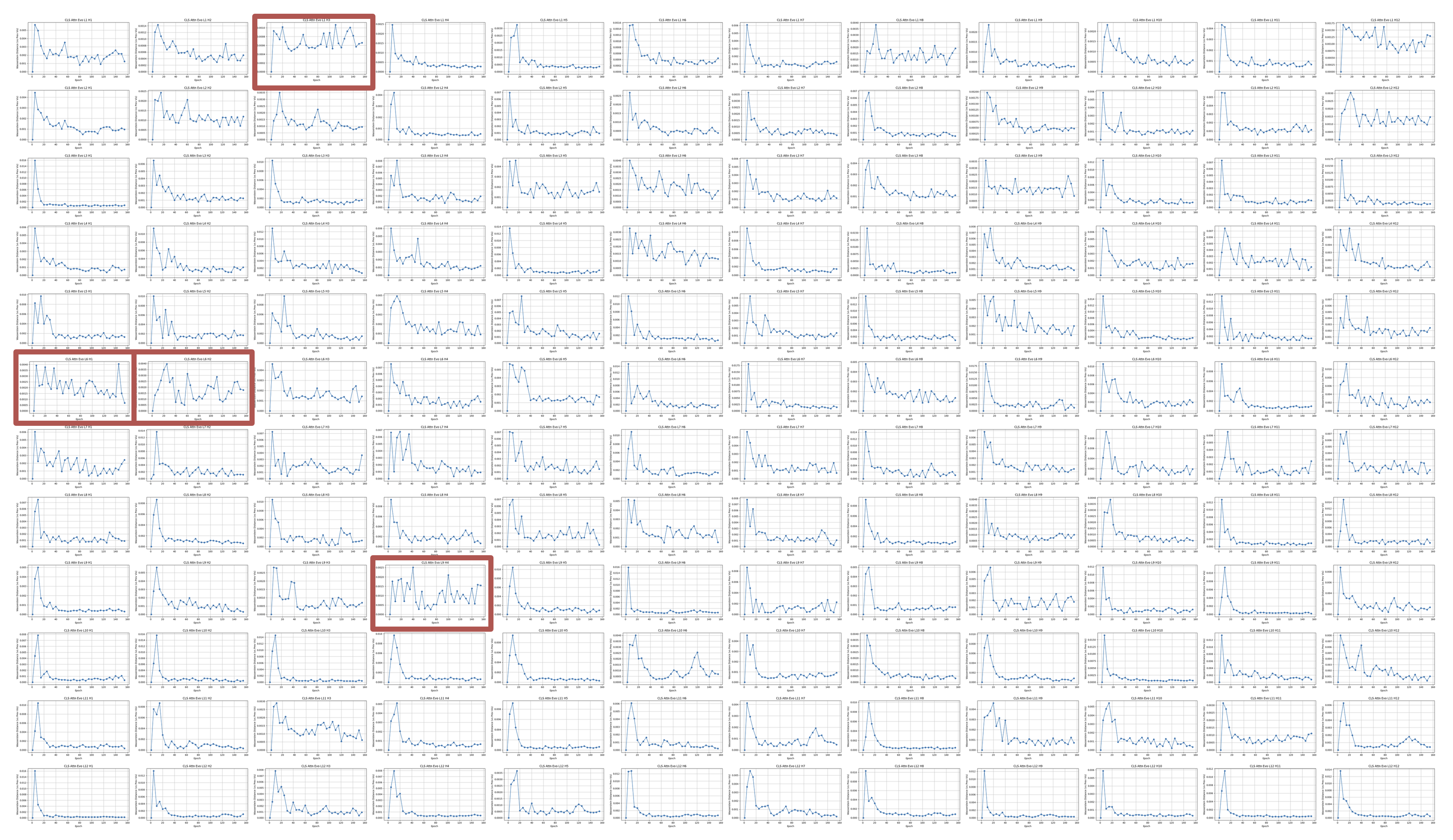}
        \caption{EMD (Attention Evolution) during training with \modelname{}'s adaptive control.}
    \end{subfigure}
    \vfill
    \begin{subfigure}[b]{\textwidth}
        \centering
        \includegraphics[width=\linewidth]{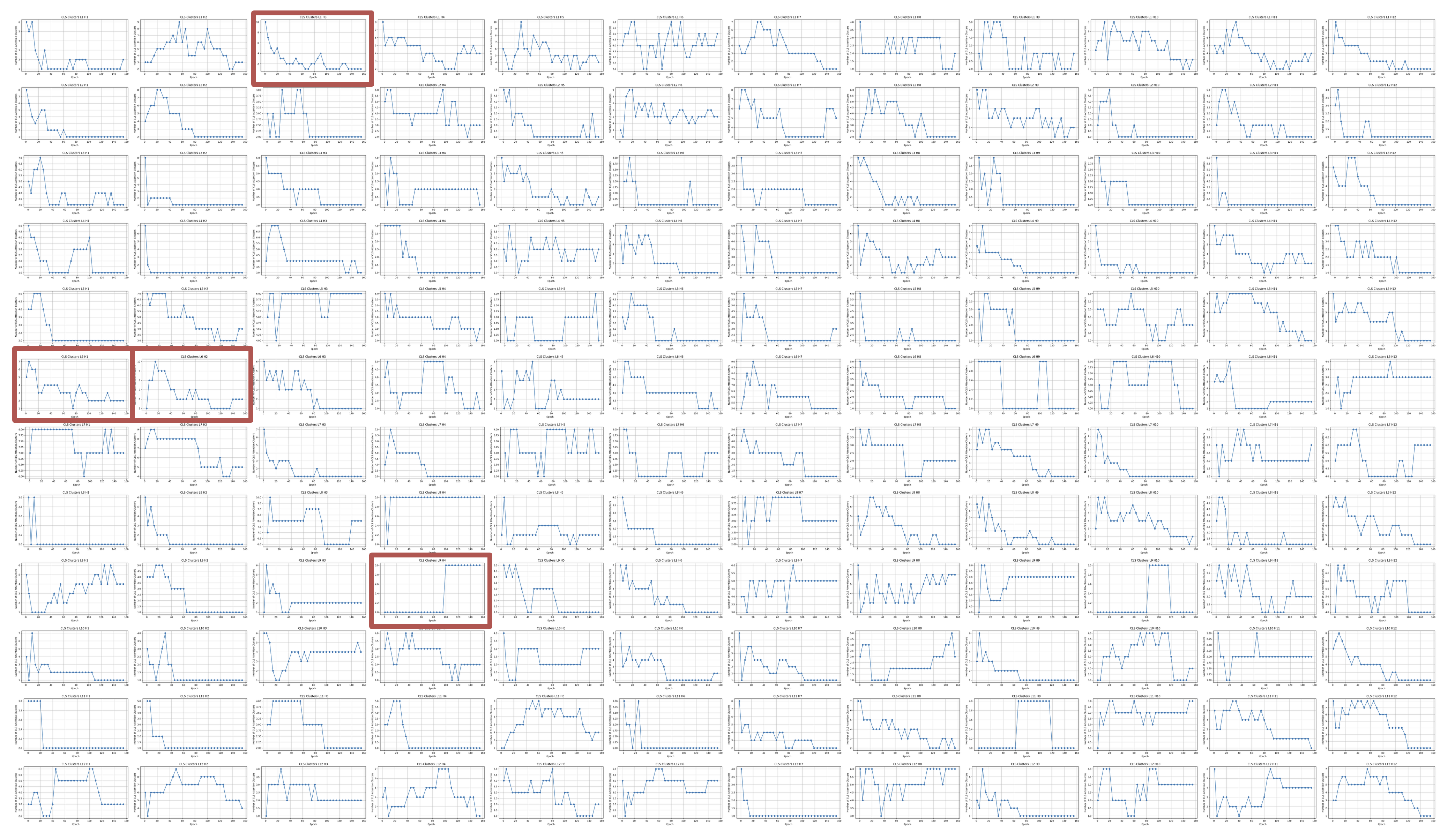}
        \caption{Number of Consensus Clusters during training with \modelname{}'s adaptive control.}
    \end{subfigure}
    \caption{\textbf{Controlled Head Dynamics.} With our adaptive temperature scheduler, EMD values for most heads rapidly decay and the number of consensus clusters stabilizes, demonstrating successful convergence of the internal dynamics.}
    \label{fig:dynamics_controlled}
\end{figure*}

In stark contrast, Figure~\ref{fig:dynamics_controlled} illustrates the effect of our adaptive control. The EMD plots for most heads show a clear and rapid decay towards zero, indicating that their attention distributions are stabilizing. This is mirrored in the consensus plots, where the number of clusters for most heads settles to a constant value early in training. This observed stability provides the empirical justification for our head-freezing mechanism; by identifying heads that have reached a stationary state, we can safely prune their computation without harming the model's learning trajectory.

\subsection{Quantitative Analysis and Training Efficiency}

Our core hypothesis is that actively controlling head dynamics improves both performance and efficiency. Table~\ref{tab:quant_results} quantifies the impact of our adaptive temperature adjustment (TA) strategy. Compared to a baseline without control, \modelname{} consistently achieves higher classification accuracy, with a notable improvement of over 7 percentage points on the CelebA dataset (89.67\% vs. 81.91\%). This performance gain is realized alongside a significant increase in computational efficiency. The number of non-converged attention heads is dramatically lower across all datasets, validating that our scheduler successfully drives heads to a stable, stationary state. This allows for effective head freezing, which reduces computational overhead in the backward pass without compromising model quality.

\begin{table}[h!]
\centering
\caption{\textbf{Quantitative comparison of control strategies.} Our Temperature Adjustment (TA) is compared against a baseline and a version with an Indicator Function (IF).}
\label{tab:quant_results}
\resizebox{0.8\linewidth}{!}{%
\setlength{\tabcolsep}{2pt} 
\small 
\begin{tabular}{@{}l l c c c@{}}
\toprule
\textbf{Dataset} & \textbf{Strategy} & \textbf{Unconv. Heads} & \textbf{Train Time (min)} & \textbf{Acc. (\%)} \\ \midrule
\addlinespace[0.3em]
Tiny ImageNet & IF + TA & 3 & 180 & 93.18 \\
Tiny ImageNet & TA (Ours) & 4 & 196 & 92.87 \\
Tiny ImageNet & Baseline & 12 & 196 & 85.33 \\ \midrule
\addlinespace[0.3em]
CIFAR-10 & IF + TA & 0 & 51 & 82.41 \\
CIFAR-10 & TA (Ours) & 0 & 60 & 82.33 \\
CIFAR-10 & Baseline & 4 & 60 & 79.01 \\ \midrule
\addlinespace[0.3em]
CelebA Face & IF + TA & 1 & 122 & 92.12 \\
CelebA Face & TA (Ours) & 3 & 138 & 89.67 \\
CelebA Face & Baseline & 8 & 138 & 81.91 \\ \bottomrule
\end{tabular}%
} 
\end{table}

\subsection{Qualitative Generative Capabilities}

\begin{figure}[h!]
\centering
\includegraphics[width=0.95\columnwidth]{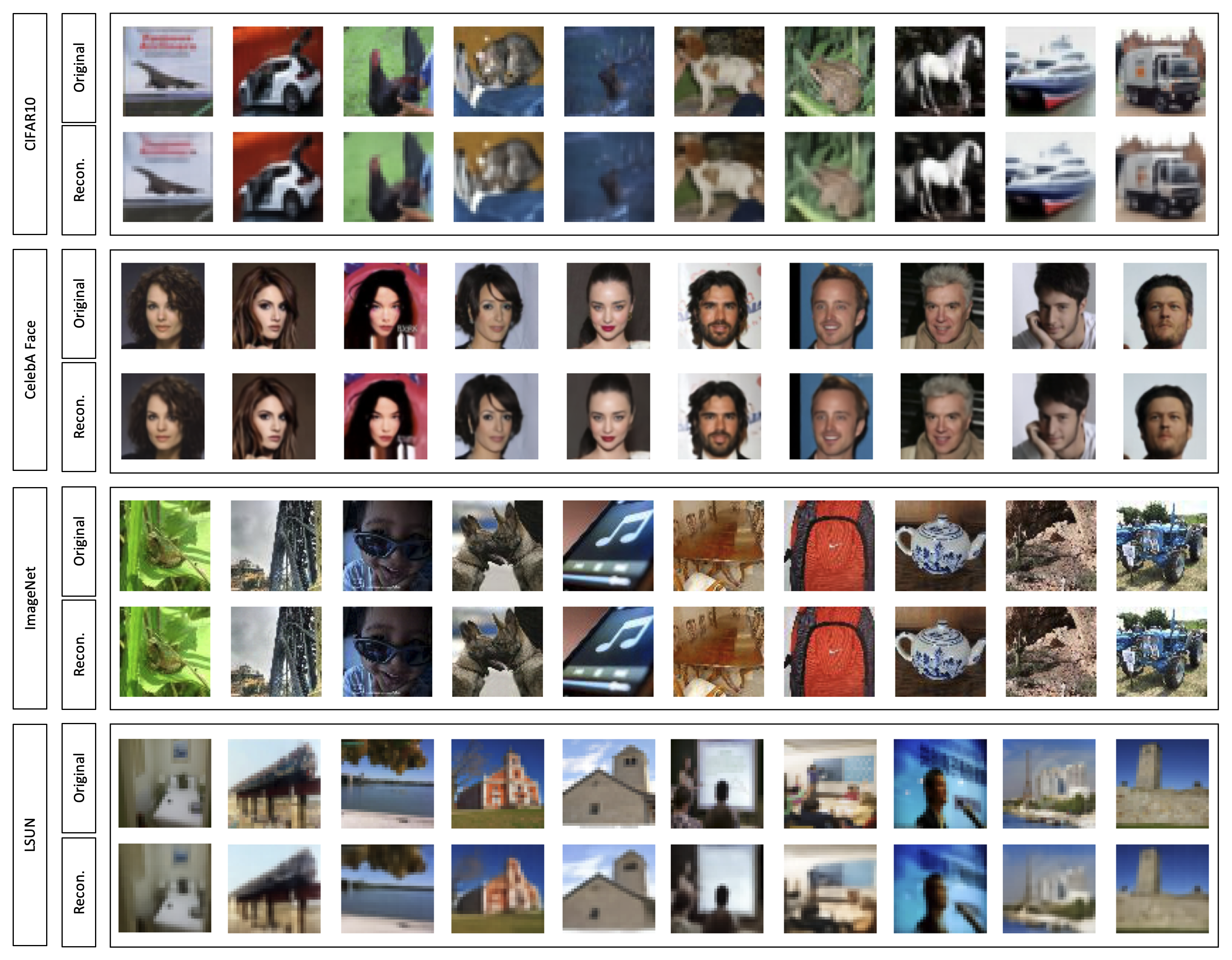}
\caption{\textbf{High-Fidelity Reconstructions.} The model accurately reconstructs input images from diverse datasets. For each dataset, the top row shows original images and the bottom row shows their corresponding reconstructions.}
\label{fig:reconstruction}
\end{figure}

We qualitatively assessed the model's ability to learn and reproduce the underlying data distribution through four distinct tasks.
\paragraph{Image Reconstruction}
As shown in Figure~\ref{fig:reconstruction}, \modelname{} produces high-fidelity reconstructions across multiple challenging datasets. The model accurately preserves both high-level semantic features and fine-grained details, demonstrating the effectiveness of the hierarchical latent structure in capturing comprehensive visual information.

\begin{figure}[h!]
  \centering
  \begin{subfigure}[h!]{0.48\columnwidth}
    \centering
    \includegraphics[width=\linewidth]{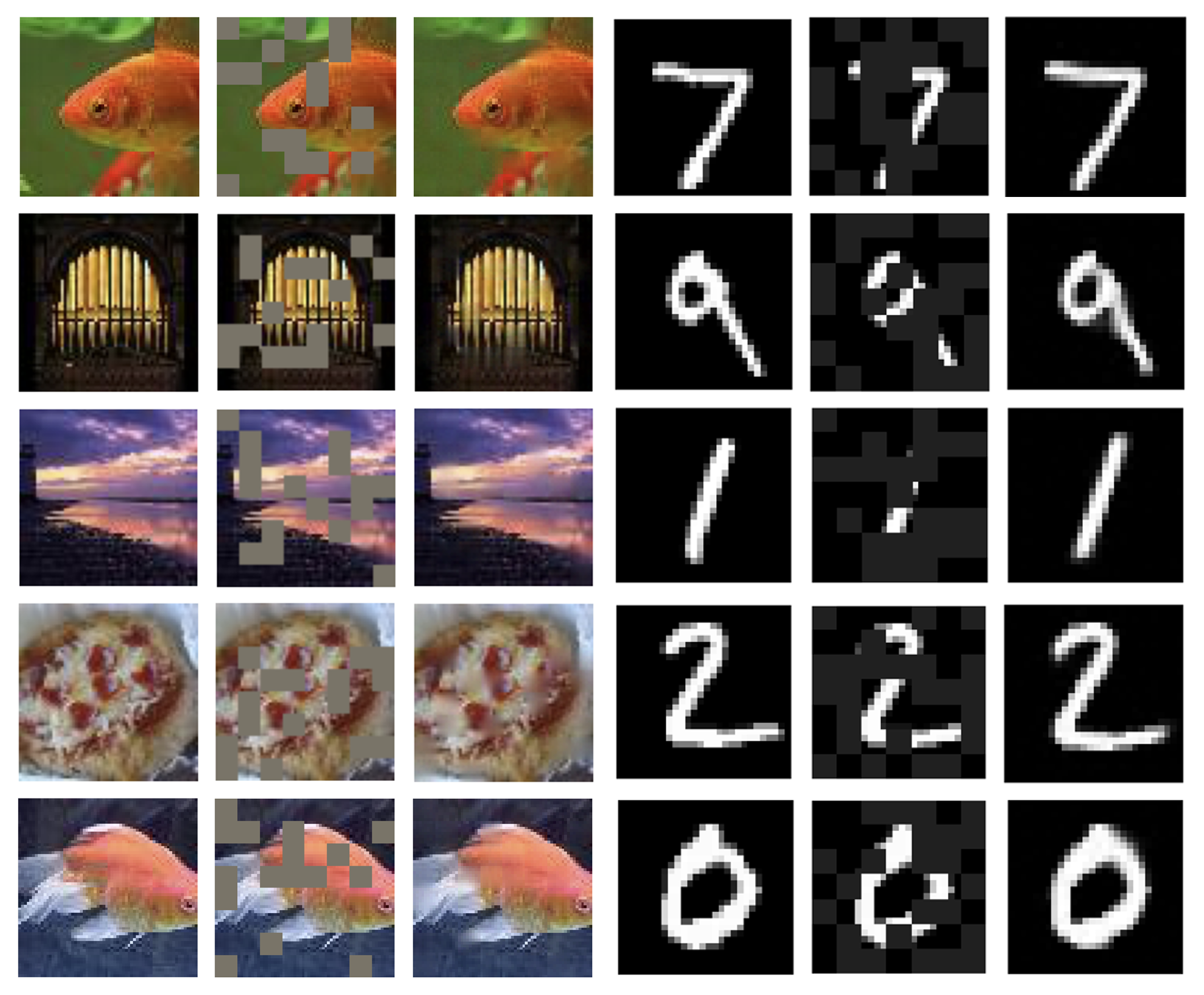}
    \subcaption{Image inpainting from masked inputs.}
    \label{fig:inpainting}
  \end{subfigure}\hfill
  \begin{subfigure}[h!]{0.48\columnwidth}
    \centering
    \includegraphics[width=\linewidth]{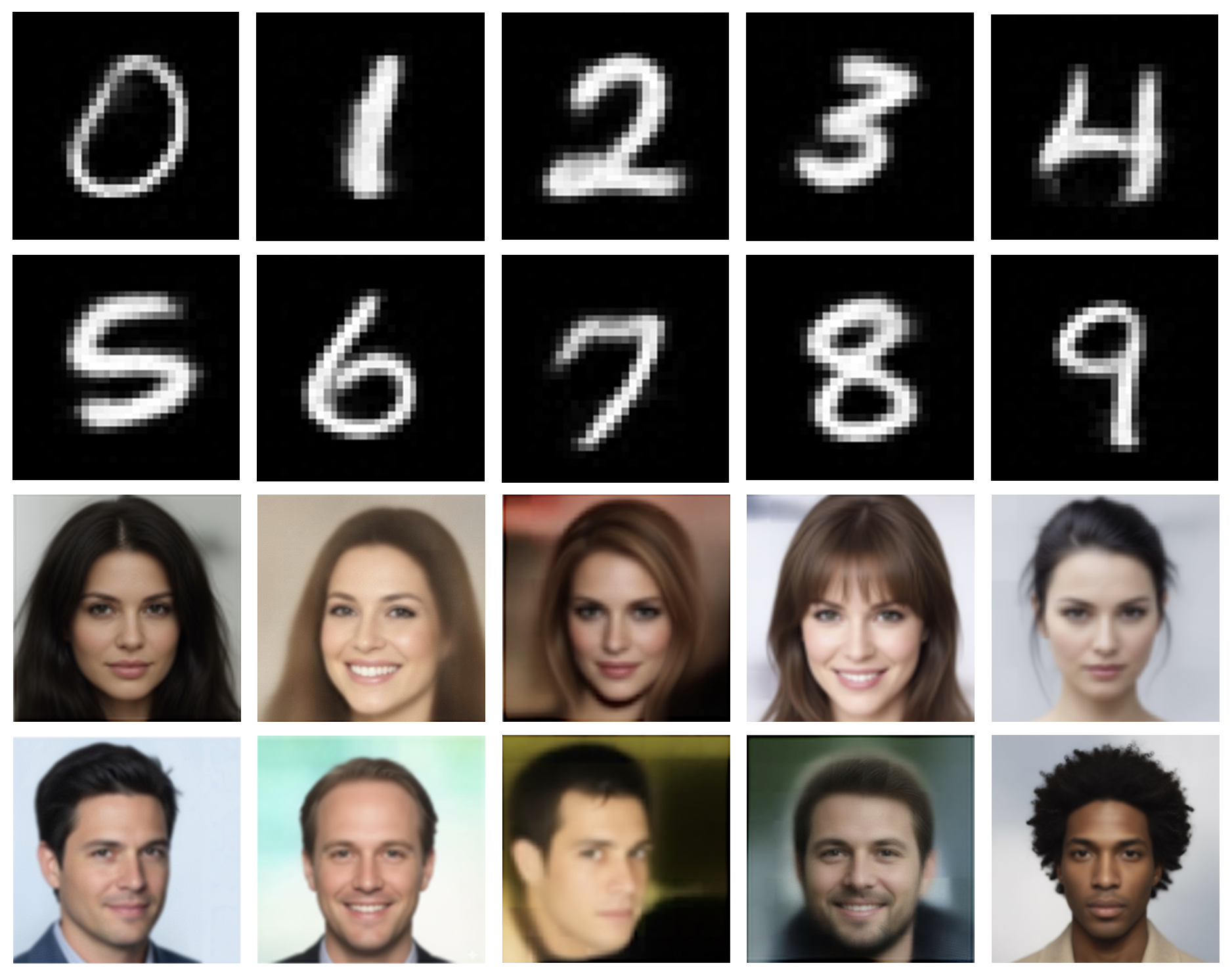}
    \subcaption{Generation out of noise.}
    \label{fig:generation}
  \end{subfigure}
  \caption{\textbf{(a) Image inpainting and (b) unconditional samples.}
  In (a), for each set, the columns show the original, masked input, and reconstruction. In (b), images generated by sampling from the prior are diverse and realistic.}
  \label{fig:inpainting-generation}
\end{figure}

\paragraph{Reconstruction from Masked Inputs}
To test the model's understanding of contextual information, we evaluated its ability to perform inpainting on heavily masked inputs. Figure~\ref{fig:inpainting} shows that \modelname{} can plausibly fill in large missing regions. The model generates coherent content that is consistent with the surrounding visible patches, indicating that it has learned meaningful spatial and semantic relationships.

\paragraph{Unconditional Generation}
By sampling from the prior distribution $p(z)$, \modelname{} can generate novel, high-quality images. Figure~\ref{fig:generation} displays samples from MNIST and CelebA. The generated images are sharp, diverse, and representative of their respective domains, confirming that the latent space has successfully captured the essential characteristics of the data distribution.

\begin{figure}[h!]
  \centering
  \includegraphics[width=0.9\columnwidth]{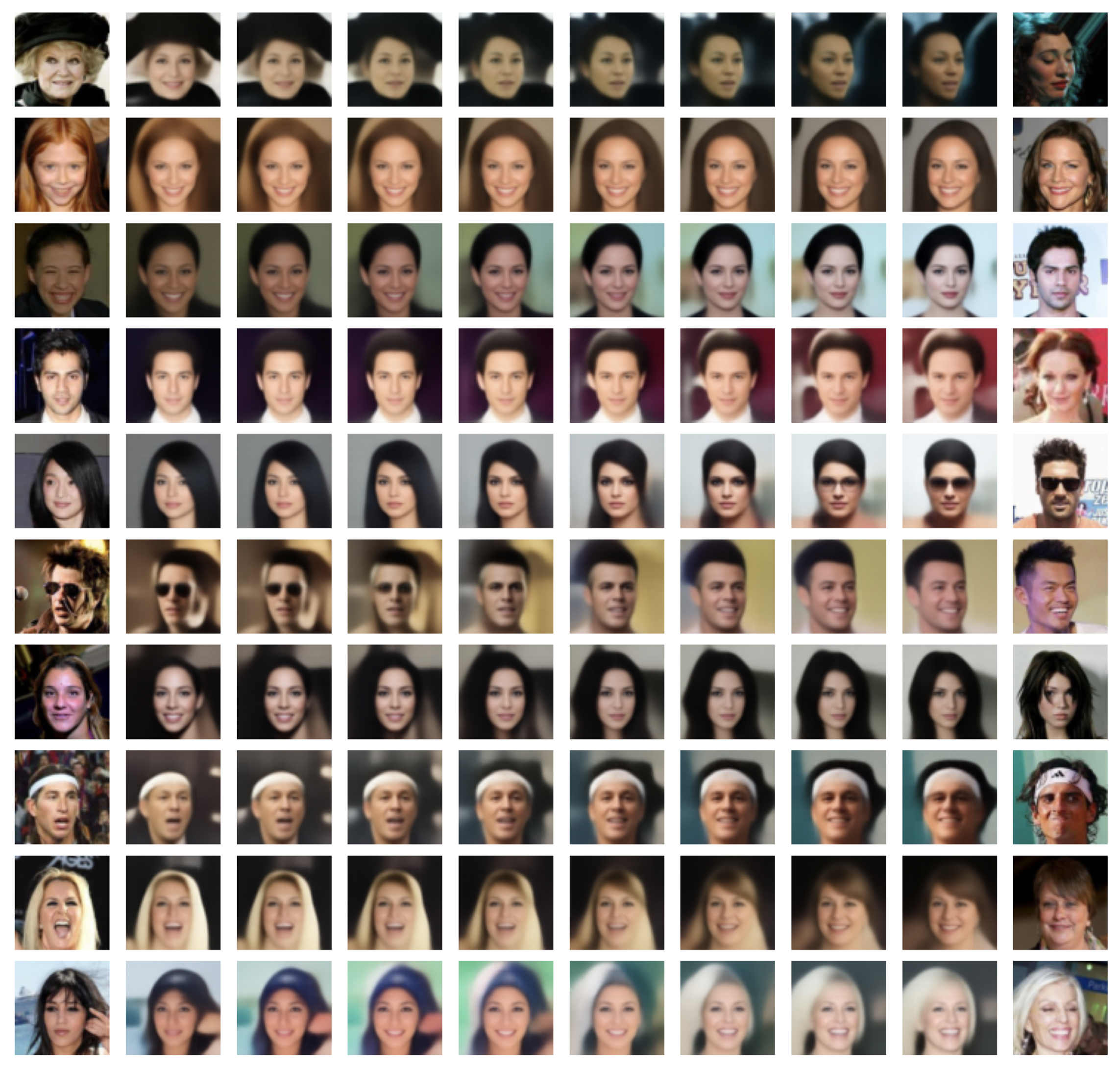}
  \caption{\textbf{Latent Space Interpolation.} Each row shows a smooth semantic transition between two real images (far left and right) by interpolating their latent vectors, highlighting a continuous latent space.}
  \label{fig:interpolation}
\end{figure}

\paragraph{Latent Space Interpolation}
To verify the smoothness and semantic structure of the latent space, we performed linear interpolation between the latent representations of two real images. As illustrated in Figure~\ref{fig:interpolation}, the model generates coherent and gradual transitions between the start and end points. Attributes like hair color and facial expression change smoothly, demonstrating that the latent space is well-organized and semantically meaningful.

\subsection{Implementation Details}
Training was performed using the AdamW optimizer. A ReduceLROnPlateau learning rate scheduler was used, monitoring validation loss, and gradient clipping (max norm 1.0) was applied. Key hyperparameters used for our experiments are summarized in Table \ref{tab:hyperparams}.

\begin{table}[h!]
\centering
\caption{Key Hyperparameters for \modelname{}}
\label{tab:hyperparams}
\resizebox{0.95\columnwidth}{!}{%
\begin{tabular}{@{}ll@{}}
\toprule
Parameter                     & Value                \\ \midrule
Batch Size                    & 512                  \\
Number of Epochs              & 200                  \\
Learning Rate (initial)       & $1 \times 10^{-4}$   \\
Image Size                    & $64 \times 64$       \\
Patch Size                    & $8 \times 8$         \\
Embedding Dimension ($D_{embed}$) & 768                  \\
Number of Attention Heads     & 12                   \\
Number of Transformer Layers ($L_{enc}, L_{dec}$)   & 12                   \\
Latent Dim \zglobal{} ($D_{z2}$)   & 256                  \\
Latent Dim \zlocal{} ($D_{z1}$)    & 64                   \\
FFN Dimension                 & $D_{embed} \times 4$ \\
PT Estimator Hidden Dim       & 1024                 \\
PT Estimator Layers           & 5                    \\
Recon L1 Coeff ($c_1$)        & 0.1                  \\
Recon L2 Coeff ($c_2$)        & 0.9                  \\
KL \zglobal{} Warmup Epochs ($K_{z2\_warmup}$)    & 5                    \\
KL \zglobal{} Weight ($w_{z2\_kl}$, post-warmup)  & $1 \times 10^{-4}$ \\
MMD \zglobal{} Weight ($w_{z2\_mmd}$, post-warmup) & $0.9999$             \\
KL \zlocal{} Warmup Epochs ($K_{z1\_warmup}$)    & 5                    \\
KL \zlocal{} Weight ($w_{z1\_kl}$, post-warmup)  & $1.0$                \\
PT Discrepancy $\lambda_{PT}$ & $0.1$                \\
MMD Kernel $C$                & 1.0                  \\
MMD Kernel $\beta_{ker}$      & 0.5                  \\
\bottomrule
\end{tabular}
}
\end{table}

\section{Conclusion}\label{sec:conclusion}

This paper introduced \modelname{}, a ViT-based VAE that repurposes the Class token into a generative linchpin to condition patch tokens, unifying global and local representations. Inspired by opinion dynamics, we proposed a dynamic training paradigm featuring a convergence-aware temperature scheduler that adaptively modulates each attention head’s influence. Guided by theoretically-grounded diagnostics, this process enables a principled head-freezing mechanism that improves computational efficiency without sacrificing performance. Our experiments confirm that this synthesis of a structured generative model with adaptive attention control yields superior results in reconstruction and generation. Future work includes scaling the architecture to higher resolutions and extending our dynamic control framework to Transformers in other domains, including language and multimodal tasks.

\newpage
\bibliographystyle{IEEEtran}

\end{document}